\documentclass[runningheads]{llncs}

\usepackage[T1]{fontenc}
\usepackage{multirow}
\usepackage[title]{appendix}
\usepackage{subfig}
\usepackage{graphicx}
\usepackage{wrapfig}
\usepackage{blindtext}
\usepackage{amsmath}
\usepackage{amssymb}
\usepackage{booktabs}
\usepackage{siunitx}
\usepackage[misc,geometry]{ifsym}
\usepackage[hidelinks,colorlinks=true,allcolors=blue,final=true]{hyperref}
\usepackage{color}

\newcommand{\etal}{\textit{et al.}}
\newcommand{\rvlcdip}{RVL-CDIP~\cite{harley2015-rvlcdip}}

\newcommand{\saliency}{Saliency~\cite{saliency}}
\newcommand{\gradcam}{GradCAM~\cite{grad-cam}}

\newcommand{\deeplift}{DeepLIFT~\cite{deeplift}}
\newcommand{\integratedgradients}{IntegratedGradients~\cite{integrated-gradients}}
\newcommand{\deepshap}{DeepSHAP~\cite{shap}}
\newcommand{\kernelshap}{KernelSHAP~\cite{shap}}
\newcommand{\lime}{LIME~\cite{lime}}
\newcommand{\occlusion}{Occlusion~\cite{occlusion}}
\newcommand{\docxplainfg}{DocXplain\textsubscript{FG}}
\newcommand{\docxplainfgbg}{DocXplain\textsubscript{FG+BG}}
\newcommand\firstval[1]{\textcolor{green}{#1}}
\newcommand\secondval[1]{\textcolor{blue}{#1}}
\newcommand\thirdval[1]{\textcolor{red}{#1}}
\makeatletter
\renewcommand\section{\@startsection{section}{1}{\z@}%
	{-8\p@ \@plus -4\p@ \@minus -4\p@}%
	{6\p@ \@plus 4\p@ \@minus 4\p@}%
	{\normalfont\large\bfseries\boldmath
		\rightskip=\z@ \@plus 8em\pretolerance=10000 }}
\renewcommand\subsection{\@startsection{subsection}{2}{\z@}%
	{-8\p@ \@plus -4\p@ \@minus -4\p@}%
	{6\p@ \@plus 4\p@ \@minus 4\p@}%
	{\normalfont\normalsize\bfseries\boldmath
		\rightskip=\z@ \@plus 8em\pretolerance=10000 }}
\renewcommand\subsubsection{\@startsection{subsubsection}{3}{\z@}%
	{-6\p@ \@plus -4\p@ \@minus -4\p@}%
	{-0.5em \@plus -0.22em \@minus -0.1em}%
	{\normalfont\normalsize\bfseries\boldmath}}

\makeatother
\begin{document}
	
	\title{DocXplain: A Novel Model-Agnostic Explainability Method for Document Image Classification\thanks{This work was supported by the BMBF projects SensAI (BMBF Grant 01IW20007)}}
	\titlerunning{DocXplain: Explainability Method for Document Image Classification}
	\author{
		Saifullah Saifullah\inst{1,2} \orcidID{0000-0003-3098-2458} \and
		Stefan Agne\inst{1,3} \orcidID{0000-0002-9697-4285} \and
		Andreas Dengel\inst{1,2} \orcidID{0000-0002-6100-8255} \and
		Sheraz Ahmed\inst{1,3} \orcidID{0000-0002-4239-6520}}
	\authorrunning{S. Saifullah \textit{et al.}}
	\institute{
		Smarte Daten and Wissensdienste (SDS), Deutsches Forschungszentrum für Künstliche Intelligenz GmbH (DFKI), Trippstadter Straße 122,
		67663 Kaiserslautern, Germany\\\email{\{firstname.lastname\}@dfki.de}\\ \and
		Department of Computer Science, RPTU Kaiserslautern-Landau, Erwin-Schrödinger-Straße 52, 67663 Kaiserslautern, Germany\and
		DeepReader GmbH, 67663 Kaiserlautern, Germany\\
	}
	\maketitle              %
	\begin{abstract}
		
		Deep learning (DL) has revolutionized the field of document image analysis, showcasing superhuman performance across a diverse set of tasks. However, the inherent black-box nature of deep learning models still presents a significant challenge to their safe and robust deployment in industry. Regrettably, while a plethora of research has been dedicated in recent years to the development of DL-powered document analysis systems, research addressing their transparency aspects has been relatively scarce. 
		In this paper, we aim to bridge this research gap by introducing DocXplain, a novel model-agnostic explainability method specifically designed for generating high interpretability feature attribution maps for the task of document image classification. In particular, our approach involves independently segmenting the foreground and background features of the documents into different document elements and then ablating these elements to assign feature importance. 
		We extensively evaluate our proposed approach in the context of document image classification, utilizing 4 different evaluation metrics, 2 widely recognized document benchmark datasets, and 10 state-of-the-art document image classification models. By conducting a thorough quantitative and qualitative analysis against 9 existing state-of-the-art attribution methods, we demonstrate the superiority of our approach in terms of both faithfulness and interpretability. 
		To the best of the authors' knowledge, this work presents the first model-agnostic attribution-based explainability method specifically tailored for document images. 
		We anticipate that our work will significantly contribute to advancing research on transparency, fairness, and robustness of document image classification models.
	\end{abstract}
	\section{Introduction}
	\label{sec:introduction}
	The recent breakthroughs in deep learning (DL) have ignited a revolution in the field of document analysis, leading to noteworthy progress in a wide variety of document understanding tasks~\cite{afzal2017-doc-class-2,ferrando2020-doc-class-4,Saifullah2022-docxclassifier,layoutlmv3,shen2021layoutparser,tilt}.
	However, the inherent lack of transparency in black-box DL models continues to raise questions about their suitability for real-world applications~\cite{safety-ml,rudin2019stop,wexler_jail_sentences}.	
	Numerous recent studies~\cite{biases-survey,geirhos2022imagenettrained,lucieri-shape-bias,hooker2020characterising,rudin2019stop} have shown that DL-based automated decision-making systems can easily succumb to data biases~\cite{biases-survey,geirhos2022imagenettrained,lucieri-shape-bias,hooker2020characterising}, which may lead to incorrect decision-making~\cite{safety-ml,rudin2019stop} or discrimination against individuals of minority groups~\cite{gender-biases,wexler_jail_sentences}. 	
	One such decision-making process is automated document image classification~\cite{afzal2017-doc-class-2,ferrando2020-doc-class-4,Saifullah2022-docxclassifier,layoutlmv3}, which may appear harmless at first but harbors the potential to become a source of great unfairness in specific scenarios---such as the unfair or biased rejection of applicant resumes from minority groups due to hidden biases present in the data.
	
	Besides the lack of transparency, deep learning models are also well-known for their vulnerability to
	minor data perturbations~\cite{intro-adversarial-2} and out-of-distribution (OOD) data~\cite{Hendrycks2019BenchmarkingNN,saifullah-robustness}. This raises questions about their robustness and underscores the necessity for an explainability framework that could be utilized to trace and validate model behavior before deployment in real-world applications.
	This requirement is especially crucial in the context of intelligent document processing, given that business documents encountered in real-world settings are not only typically corrupted with novel degradations~\cite{saifullah-robustness,ShabbyPages2023} but also frequently appear in various novel layouts, featuring content that was unseen during the training phase~\cite{harley2015-rvlcdip,li2020docbank,doclaynet}.
	Such OOD data samples may lead to failures at test time, making the traceability of decisions important for determining the reasons behind such failures.
	
	The field of eXplainable AI (XAI)~\cite{intro-survey-1,intro-survey-2} has recently emerged to address the aforementioned issues of transparency and accountability, with numerous methods proposed in recent years~\cite{lime,grad-cam,smooth-grad,shap,saliency,nemirovsky2020countergan} that attempt to explain the predictions of black-box DL models. 
	However, despite significant attention devoted to the development of explainability methods for natural images, research specifically tailored for document images has been relatively scarce~\cite{Saifullah2023-docxai,Saifullah2022-docxclassifier}.  
	Moreover, in a recent benchmark study, Saifullah~\etal~\cite{Saifullah2023-docxai} demonstrated that applying existing attribution-based XAI methods, commonly used in the natural image domain, to the document images presents several challenges, such as extensive noise in the output attributions, difficulty in interpreting explanations, and significant variation in explanations generated by different methods. In addition, the binary nature of document images poses a challenge in assessing whether a model emphasizes textual information in the foreground or comprehends the overall structural information of the document to arrive at its decision.	
	
	In this paper, we attempt to tackle the aforementioned challenges and propose a novel algorithm for generating explanations that takes into account the binary nature of the document images. Our approach mainly draws inspiration from the model-agnostic perturbation-based approaches~\cite{occlusion,lime,shap}, which assign importance to different image features by perturbing those features and evaluating the perturbations' impact on the model's decision. 
	However, it has been demonstrated that standard perturbation-based approaches~\cite{occlusion,lime,shap} behave poorly when applied directly to document images~\cite{Saifullah2023-docxai}, mainly due to the indiscriminate perturbation of image patches without regard for the structure in the data. To address this, we present an approach to independently segment the foreground and background features of the documents into different structural elements and at different levels of granularity, regardless of the document type. Then, feature ablation is applied to the segmented regions to allocate per-pixel importance, resulting in an image attribution map.
	Overall, the contributions of this paper can be summarized as follows:
	\begin{itemize}
		\item[--] We introduce DocXplain, a model-agnostic explanation framework especially designed to generate fine-grained attribution maps for the task of document image classification. To the best of the authors' knowledge, this is the first model-agnostic explanation framework specifically tailored for document images.		
		\item[--] We evaluate our proposed framework on the task of document image classification with 4 different evaluation metrics, 2 document datasets, and 10 distinct deep learning (DL) models, and compare the results with 9 existing state-of-the-art attribution methods.	
		Through an exhaustive quantitative and qualitative analysis, we demonstrate that our approach either outperforms or performs comparably to existing state-of-the-art attribution-based approaches.
	\end{itemize}
	
	\section{Related Work}
	\label{sec:related-work}
	\subsection{eXplainable AI (XAI)}
	The field of XAI has been extensively explored in the past decade~\cite{intro-survey-1,intro-survey-2}, resulting in a wide variety of approaches~\cite{saliency,input-x-gradient,integrated-gradients,shap,lime} for generating explanations regarding the predictions of black-box neural networks. In the following sections, we present a brief overview of the existing XAI methods.
	
	\subsubsection{Feature attribution methods.} 
	Feature attribution methods assign importance to input features or feature groups by locally estimating their relevance to the model's decision, and they are particularly recognized for their post-hoc applicability.
	These methods can be broadly divided into two main categories: gradient-based approaches and perturbation-based approaches. 
	Gradient-based methods such as \saliency{}, \gradcam{}, \deeplift{}, and \integratedgradients{} utilize gradient back-propagation inherent to black-box neural networks and assign importance to input features by approximating their contribution to the model gradients. 
	In contrast, perturbation-based approaches, such as \occlusion{} and Local Interpretable Model-Agnostic Explanations (LIME)~\cite{lime}, generate importance maps by perturbing the input features and evaluating the impact of perturbation on the model's prediction, either directly or by estimating it through surrogate machine learning models.
	
	SHapley Additive exPlanations (SHAP) is another popular group of feature-attribution approaches~\cite{shap} that utilizes Shapley values---a game theory concept---to model feature importance. In this framework, each input feature is akin to a \textit{player} in a game, and the output of the model is treated as the \textit{payout}, which corresponds to the importance of the input features. 
	SHAP~\cite{shap} has several extensions, each tailored to explain different types of machine learning models. For instance, TreeSHAP~\cite{lundberg2019consistent} is designed to explain the outputs of random forests through Shapley values. \kernelshap{} is a model-agnostic explanation framework that extends \lime{} and computes Shapley values by training surrogate models on perturbed inputs. Similarly, \deepshap{} extends the gradient-based approach of \deeplift{} and utilizes the gradients of the model to generate the Shapley values for assigning feature importance.
	
	\subsubsection{Concept-based approaches.} Concept-based approaches utilize predefined human-interpretable concepts to provide global explanations for machine learning models. The most prominent approach in this domain is Testing with Concept Activation Vectors (TCAV)~\cite{tcav} that utilizes concepts, such as stripes and textures in the case of natural images, to determine whether a neural network has learned relevant features from the data.
	
	\subsubsection{Generative approaches.} Generative approaches have also gained considerable attention in recent years, notably due to their ability to generate high-quality and intuitive counterfactual explanations~\cite{lang2021explaining,jeanneret2022diffusion}. These approaches typically employ external generative models, such as generative adversarial networks (GANs)~\cite{goodfellow_generative_2014} or diffusion models~\cite{ddpm}, to synthesize features based on their relevance to the model's prediction.
	
	\subsection{XAI in Document Image Analysis}
	While a plethora of research has gone into developing XAI methods for natural images,  there has been relatively little exploration of XAI in the domain of document image analysis. 
	The earliest work in this direction was conducted by Tensmeyer~\etal~(2017)~\cite{tensmeyer2017-doc-class-3}, who investigated the intermediate layer activation maps to explain the decision-making of DL-powered document image classification models. 
	Brini~\etal~\cite{explain_doc_seg} recently presented an end-to-end explanation framework for document layout analysis.  Their approach, however, simply re-uses existing gradient-based methods~\cite{grad-cam,deeplift} to generate the explanations for pre-annotated segmented regions.
	Saifullah~\etal~\cite{Saifullah2023-docxai} presented an extensive benchmark of several existing state-of-the-art feature-attribution methods for the task of document image classification. In this work, they raised several concerns regarding the interpretability and fidelity of existing methods when applied to the document domain, noting that only a few techniques yielded sufficiently viable results. In another recent work, Saifullah~\etal~\cite{Saifullah2022-docxclassifier} presented DocXClassifier, an inherently interpretable document image classification model that combines a convolutional neural network (CNN) with attention-based pooling mechanism to produce attribution maps at inference. However, this method has some limitations of its own: (1) the model is only capable of generating coarse attribution maps, unable to highlight fine-grained features in the image, and (2) the approach is not model-agnostic, requiring modifications to existing neural network architectures and additional training.

	\section{DocXplain: Proposed Approach}		
	In this section, we present our proposed explanation framework for generating fine-grained attribution maps for the task of document image classification. As illustrated in Fig.~\ref{fig:approach}, our approach mainly comprises two stages: (1) Feature Segmentation and (2) Feature Ablation, both of which will be explained in detail in the following sections.	
	
	\subsection{Feature Segmentation}
	The first step in our approach involves segmenting the images into distinct features. Since we are dealing with document images, the features in
	this case correspond to both the black pixels representing the textual, tabular, or imagery content in the foreground and the white pixels representing the background.	
	Note that various existing methods, including DL-based segmentation or optical-character-recognition (OCR) models, could have been utilized for segmenting the document images into relevant regions.
	However, while investigating these approaches, we noted that they are generally data-driven and capable of handling only specific types of inputs. For instance, an OCR model may perform poorly on handwritten text, making it unsuitable for generating reasonable segmentation maps for handwritten text documents. Therefore, to devise a more generic segmentation approach applicable to many different types of input documents, we opted to build it upon traditional image processing techniques.
	Formally, let $I\in \mathbb{R}^{C\times H\times W}$ denote a full-page input document image.
	Then, our feature segmentation approach involves the following steps:
	
	\subsubsection{Preprocessing.}
	The first step in our approach involves converting the image into a single-channel binary image ($I\in \mathbb{R}^{H\times W}$) with black (foreground) and white (background) pixels. This can be easily achieved using standard image processing algorithms such as Otsu binarization~\cite{Otsu1979ATS}. It is important to note that here we assume sufficiently clean input document images (gray-scale or RGB) that can be effectively binarized into distinct background and foreground regions, which is a common scenario with business documents.
	In the second step of our approach, we downscale the images to a fixed resolution of $1024\times 1024$ without interpolation (to maintain the image in binary format) and apply the standard opening operation (dilation followed by erosion) with a $5 \times 5$ filter to remove noise from the image. 
	
	\subsubsection{Background Segmentation.}
	While the foreground features, characterized by black pixels, are generally distinguishable in document images, the background simply consists of white pixels with no internal structure. Therefore, to segment the background, we generate a grid over the image with a fixed patch size of $P_{bx}\times P_{by}$. This results in the background segmentation matrix $M_{bg} : \mathbb{X} \rightarrow \{1,\dots,n_{bg}\}$, $\mathbb{X} = \{1,...,H\}\times\{1,...,W\}$ with a total of $n_{bg}=\frac{W}{P_{bx}}.\frac{H}{P_{by}}$ segmented regions over the whole image $I\in \mathbb{R}^{H\times W}$.
	
	\begin{figure}[t!]
		\includegraphics[width=\textwidth]{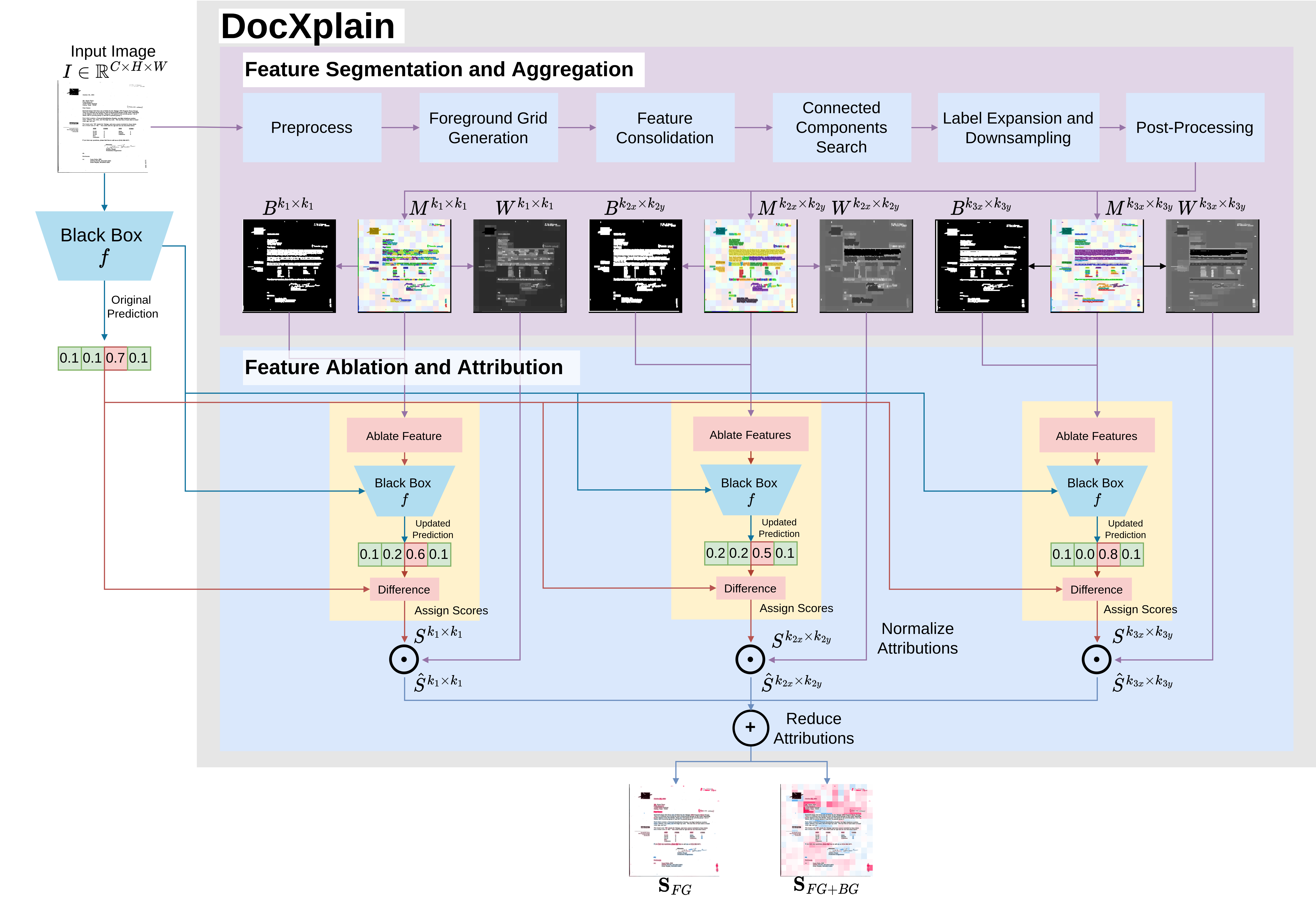}
		\caption{An overview of the proposed approach, DocXplain, is presented. As shown, the image first undergoes several processing steps for generating segmentation maps at different kernel sizes. Then, given any black-box neural network $f$, per-kernel feature importance maps are generated by applying feature ablation in combination with the respective segmentation maps. Subsequently, the maps are summed to simultaneously create decoupled attribution maps for both background and foreground regions. $\mathbf{S}_{FG}$ and $\mathbf{S}_{FG+BG}$  correspond to the \docxplainfg{} \docxplainfgbg{} settings, respectively.} \label{fig:approach}
		\vspace{-1.5em}
	\end{figure}
	\subsubsection{Foreground Segmentation.}
	To segment the foreground regions, we first apply the standard dilation operation with a fixed kernel size of $k\times k$ on the inverted image in order to consolidate foreground (text or imagery) regions into blobs. 
	Then, we apply connected components search on the image to assign a distinct label to each foreground blob, whereas all the white background pixels are assigned a fixed label of $0$ at this stage. This yields a segmentation mask $M_{fg} : \mathbb{X} \rightarrow \{0,\dots,n_{fg}\}$, 
	where $n_{fg}$ represents the total number of foreground features extracted from the image.
	Note that depending on the specific requirements, different kernel sizes $k_1\times k_2$ may be used at this stage to obtain feature segmentation at different resolutions.
	Finally, we combine the two segmentation masks $M_{bg}$ and $M_{fg}$ by taking the following steps. We first offset the unique labels of $M_{fg}$ by $n_{bg}$ to generate a new foreground mask with the labels in the range $\{0, 1+n_{bg},\dots,n_{fg}+n_{bg}\}$. Notice that the offset is applied only to the foreground labels in $M_{fg}$, while the background label $0$ remains unchanged. The combined segmentation mask $M : \mathbb{X} \rightarrow \{1,\dots,n_{bg}+n_{fg}\}$ is then computed as follows:
	\begin{equation}
	M(\lambda\in \mathbb{X}) = \begin{cases}
	M_{bg}(\lambda) &\text{if $M_{fg}(\lambda) = 0$ (background labels)}
	\\
	M_{fg}(\lambda), &\text{else (foreground labels)}
	\end{cases}
	\end{equation}	
	\subsubsection{Post Processing.}
	After generating the initial segmentation map, we perform label expansion on the foreground elements (where $l > n_{bg}$) for $N_{expansion}$ pixels and then downsample the image to a resolution required at the model input, such as $224 \times 224$, which is commonly utilized for deep-learning models. Label expansion at this step was necessary, as otherwise, downsampling an image to very small resolutions can result in the loss of important information about segmentation boundaries. For preprocessed images of resolution $1024\times 1024$, we found that expanding the labels by $2$ pixels ($N_{expansion}=2$) was suitable. In addition, downsampling was necessary at this stage to reduce the computational overhead for further processing. Despite these adjustments, two major issues remained:
	\begin{enumerate}
		\item There may be connected components left that span the entire image while occupying a very small area, such as tables or graphs. On the other hand, large figures or images may be represented as a single connected component while containing specific distinct features within.
		\item The segmentation process may result in an extensively large number of small-sized noisy connected features, which can be computationally infeasible to process in later stages.
	\end{enumerate}
	To address the first problem concerning large components such as  tables, images, or figures, we utilize the Simple Linear Iterative Clustering (SLIC)~\cite{slic} algorithm to further cluster these components into SLIC superpixels. As for the second problem, we mitigate it by performing a second stage of denoising, in which we substitute the labels of features with areas smaller than a fixed threshold with the corresponding background labels.	
	\begin{figure}[t!]
		\includegraphics[width=\textwidth]{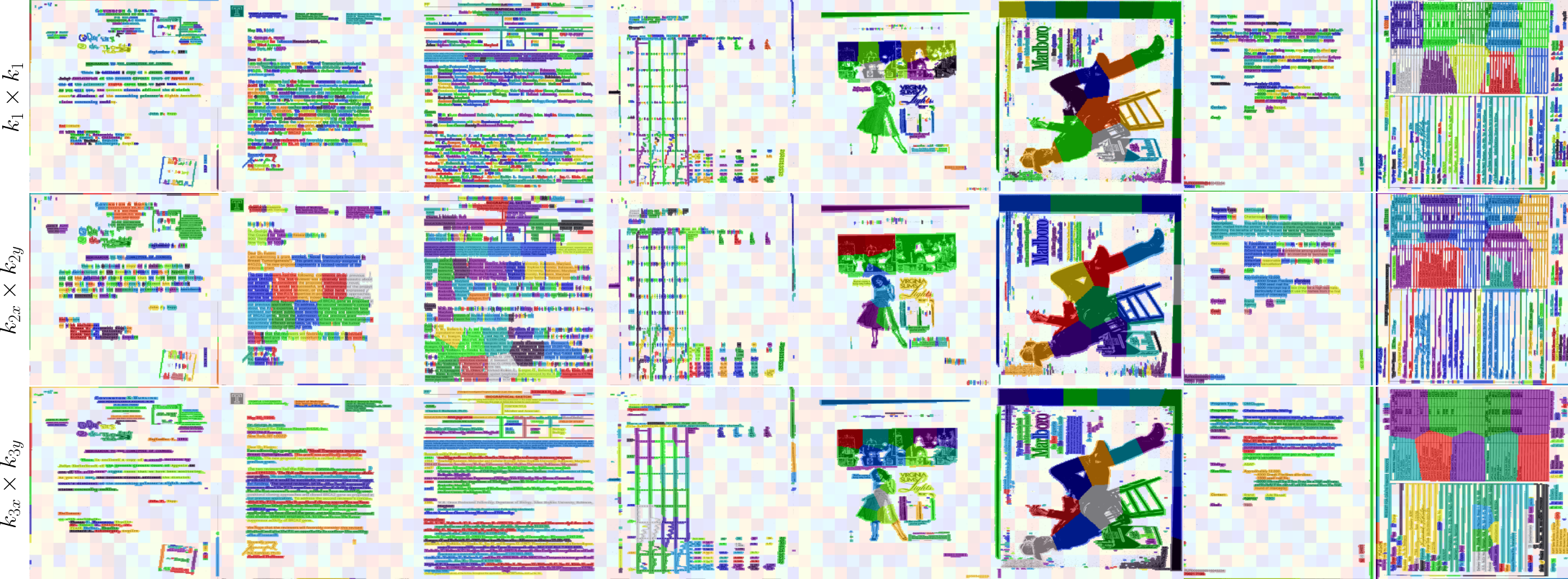}
		\caption{Segmentation maps generated by our approach for the kernels $k_1\times k_1=5 \times 5$, $k_{2x}\times k_{2y}= 3 \times 15$, and $k_{3x}\times k_{3y} = 15 \times 3$ on a few randomly selected samples from the Tobacco3482 document dataset. } \label{fig:kernel_sizes}
		\vspace{-0.5em}
	\end{figure}
	\subsubsection{Choice of Segmentation Kernels.}	
	Since document images are semi-structured, it is challenging to assess whether, for a model's prediction, an individual word, a text line, or a text block, and so forth, was found relevant. 
	Therefore, in this work, we use 3 different sizes of kernels to segment the images at 3 different levels of granularity:
	\begin{itemize}
		\item A kernel of small size $k_1\times k_1$, to generate segmentation maps for each individual textual content in the image.
		\item A kernel of size $k_{2x}\times k_{2y}$, such that $k_{2y} \gg k_{2x}$, to consolidate regions that are vertically close in distance, such as individual text blocks or a single column in tabular data.		
		\item A kernel of size $k_{3x}\times k_{3y}$,  where $k_{3x} \gg k_{3y}$, to consolidate regions that are horizontally aligned such as text lines.
	\end{itemize}
	Note that these kernel sizes can be considered hyperparameters and were chosen empirically for the datasets under consideration. In our experiments, we found that the configuration $k_1\times k_1=5\times 5$, $k_{2x}\times k_{2y}=3\times 15$, and $k_{3x}\times  k_{3y}=15\times 3$, worked the best.	Fig.~\ref{fig:kernel_sizes} illustrates the segmentation maps resulting from our approach using these three kernel sizes on a few randomly selected samples from the Tobacco3482 document dataset.
	
	\subsection{Feature Ablation}
	Let $f: \mathcal{I} \rightarrow \mathbb{R}$ denote a black-box deep neural network to be explained, that for a given input image from the $\mathcal{I}=\{I \mid I:\mathbb{X} \rightarrow R^C\}$, $\mathbb{X} = \{1,...,H\}\times\{1,...,W\}$ with size $H \times W$ and number of channels $C$, produces a scalar output confidence score. 
	For the task of document image classification, the output of $f$ is typically the prediction probability score for a given document class. 
	Then, given the segmentation mask $M : \mathbb{X} \rightarrow \{1,\dots,n_{bg}+n_{fg}\}$, we define the importance of each feature group $M_i =\{\lambda\in \mathbb{X} \mid M(\lambda)=i\}, i\in\{1,\dots,n_{bg}+n_{fg}\}$ as the difference in confidence score observed when that feature is removed from the image $I$:
	\begin{equation}
	S_i = f(I) - f(I \setminus M_i) \label{eq:ablation}
	\end{equation} 
	where the term $f(I \setminus M_i)$ represents the removal of feature $M_i$ from the image $I$. Since document images are generally processed in gray-scale or binary format, we perform feature removal by substituting the pixel values of each foreground element with the background pixels and vice versa. More formally, for an image with pixel values normalized between 0 and 1, we use the following baseline matrix $B:\mathbb{X} \rightarrow \{0, 1\}^C$ to replace the pixels of each feature group $M_i$:
	\begin{equation}
	B(\lambda\in \mathbb{X}) = \begin{cases}
	0 &\text{if $M(\lambda) \leq n_{bg}$ (background regions)}
	\\
	1, &\text{else (foreground regions)}
	\end{cases}
	\end{equation}
	Applying Eq.~\ref{eq:ablation} on the model for all feature groups $i\in \{1,\dots,n_{bg}+n_{fg}\}$ results in the output attribution score matrix $S:\mathbb{X} \rightarrow \mathbb{R}$, where $S(\lambda\in \mathbb{X})=\{S_i \mid M(\lambda)=i\}, i\in\{1,\dots,n_{bg}+n_{fg}\}$. 
	Note that at this stage, since the attribution scores $S$ are computed by removing grouped features in each iteration, the resulting scores are unnormalized with respect to the group size.	
	For instance, removing larger feature groups will inherently result in larger drops in the model confidence as compared to the removal of smaller groups. To account for this bias, we normalize the attribution score of each feature set $M_i$ by the inverse of the area occupied by it, and compute  the final attribution score matrix as follows:
	\begin{equation}
	\hat{S} = S \odot \mathcal{W}
	\end{equation} 
	where $\mathcal{W}:\mathbb{X} \rightarrow \mathbb{R}$ is the weight matrix defined as $\mathcal{W}(\lambda \in \mathbb{X}) = \frac{1}{|\{\lambda\in \mathbb{X}\mid M(\lambda)=i\}|}=\frac{1}{|M_i|}, i\in\{1,\dots,n_{bg}+n_{fg}\}$ and $\odot$ denotes the element-wise multiplication.
	Finally, with $K$ different segmentation masks $M^{k}, k\in\{1,\dots,K\}$, the final attribution matrix $\textbf{S}: \mathbb{X} \rightarrow \mathbb{R}$ over each image can be simply computed as the sum of the normalized score matrices $\hat{S}^{k}$ of each mask $M^{k}$:
	\begin{equation}
	\textbf{S} = \sum_{k}^{K}\hat{S}^k \label{eq:final-attr}
	\end{equation} 
	With our choice of kernels to generate 3 different segmentation maps, Eq.~\ref{eq:final-attr} can be rewritten as follows:
	\begin{equation}
	\textbf{S} = \hat{S}^{k_1\times k_1} + \hat{S}^{k_{2x}\times k_{2y}} + \hat{S}^{k_{3x}\times k{3y}}
	\end{equation} 
	In all our experiments, we evaluate our approach under two settings: \docxplainfg{} and \docxplainfgbg{}. In \docxplainfg{}, only the foreground segmented features are ablated, whereas, in \docxplainfgbg{}, both foreground and background features are ablated, as illustrated in Fig.~\ref{fig:approach}.
	
	\section{Experiment Setup}
	\subsection{Datasets} 
	We thoroughly investigate the effectiveness of our approach on two of the most widely used document image classification benchmarks: \rvlcdip{} and Tobacco3482\footnote{\url{https://www.kaggle.com/datasets/patrickaudriaz/tobacco3482jpg}}. \rvlcdip{} is a large-scale document image dataset with 16 equally distributed document classes and has been widely utilized as a benchmark for the task of document image classification~\cite{afzal2015-breakthrough-tl,afzal2017-doc-class-2,harley2015-rvlcdip,ferrando2020-doc-class-4,Saifullah2022-docxclassifier,layoutlmv3,tilt}. The dataset showcases significant sample diversity, consisting of a total of 400,000 annotated document images with training, testing, and validation splits of sizes, 320,000, 40,000 and 40,000, respectively. The Tobacco3482$^{1}$ dataset, on the other hand, is small-scale, with only 3482 document samples divided into 10 different document types. In contrast to \rvlcdip{}, however, this dataset has a significant class imbalance. In our experiments, we split this dataset into training, testing, and validation sets of sizes 2,504, 700, and 279, respectively.
	
	\subsection{Models}	
	To evaluate the generalizability of our proposed approach across different model architectures, we choose 10 different deep learning models, namely, AlexNet~\cite{alexnet}, ResNet-50~\cite{resnet}, InceptionV3~\cite{inception}, VGG-16~\cite{vgg}, MobileNetV3~\cite{mobilenet}, NFNet-F1~\cite{nfnet}, DenseNet-121~\cite{densenet}, Res2Net-50\cite{res2net}, EfficientNet-B4~\cite{effnet}, and ConvNeXt-B~\cite{convnext}. Note that the same set of models was also previously investigated in the benchmark study by Saifullah~\etal~\cite{Saifullah2023-docxai} and therefore provides a good comparison with the previous work. To train the models on the aforementioned datasets, we initialized them with ImageNet pretraining weights, first fine-tuned them on the \rvlcdip{} dataset, and then further fine-tuned them on the Tobacco3482 dataset. In particular, the first 8 models were trained using the approach proposed by Afzal~\etal~(2017)~\cite{afzal2017-doc-class-2}, whereas the remaining two models, EfficientNet-B4~\cite{effnet} and ConvNeXt-B~\cite{convnext}, were trained using the methodologies outlined in the original works by Ferrando~\etal~\cite{ferrando2020-doc-class-4} and Saifullah~\etal~\cite{Saifullah2022-docxclassifier}, respectively, in which these models were proposed for the task of document image classification. 
	Following previous works~\cite{afzal2017-doc-class-2,ferrando2020-doc-class-4,Saifullah2022-docxclassifier}, for all models, except EfficientNet-B4 and ConvNeXt-B, we utilized input image resolutions of $224\times224$, while for these two models, we used input image resolutions of $384\times384$. The performance achieved by the models on the respective test sets is given in Table.~\ref{tab:acc}.
	
	\begin{table}[t!]
		\centering
		\caption{The performance of each model on the test sets of \rvlcdip{} and Tobacco3482 is listed. All the models were trained to achieve sufficiently high performance on the respective datasets for valid explainability analysis.}\label{tab:acc}
		\setlength{\tabcolsep}{0.5em}
		\resizebox{\textwidth}{!}{		
			\begin{tabular}{lcccccccccc}\toprule
				&\multicolumn{10}{c}{Models}\\
				\cmidrule{2-11}
				Datasets           & AlexNet & ResNet50 & InceptionV3 & VGG-16 & MobileNetV3 & NFNet-F1 & DenseNet121 & Res2Net50 & EfficientNet-B4 & ConvNeXt \\ \midrule
				RVL-CDIP (Acc \%) & 87.86   & 90.40    & 91.18       & 90.96  & 87.56       & 88.83    & 89.46       & 88.96     & 92.68            & \textbf{93.89} \\
				Tobacco3482 (Acc \%) & 88.85 & 92.00    & 92.71       & 93.71  & 85.85       & 91.28    & 92.14       & 89.85     & 93.85            & \textbf{94.71} \\ \bottomrule
			\end{tabular}
		}
	\vspace{-0.5em}
	\end{table}
	\subsection{XAI Methods}
	We compare our approach with 9 different existing attribution-based methods, which include Saliency~\cite{saliency}, GuidedBackprop~\cite{guided-backprop}, IntegratedGradients~\cite{integrated-gradients}, InputXGradient~\cite{input-x-gradient}, and DeepLIFT~\cite{deeplift}, SHAP-based methods including DeepSHAP \cite{shap} and KernelSHAP~\cite{shap}, and perturbation-based methods including LIME~\cite{lime} and Occlusion~\cite{occlusion}. For all the gradient-based methods, we use default parameters for computing attributions. For segmentation-based perturbation-based approaches \kernelshap{} and \lime{}, we directly apply the SLIC~\cite{slic} algorithm to generate superpixels as proposed in \lime{} and then train the surrogate models. 
	For \deepshap{}, we utilize a background distribution of 100 samples for each sample, which we found sufficient for computing the attributions. Note that increasing the number of samples in this background distribution may enhance performance, but it also leads to extensively high computational requirements, which we found infeasible for our experiments.	
	Finally, we apply \occlusion{} with a patch size of $16\times 16$ and a stride of $8\times 8$ for images of size $224\times224$ and a patch size of $32\times 32$ and a stride of $16\times 16$ for images of size $384\times384$. Again, it is worth mentioning that decreasing the patch size in \occlusion{} may result in more fine-grained attributions but at huge computational costs.
	
	\subsection{Evaluation Metrics}
	\label{sec:metrics}
	We perform a quantitative comparison of our approach with existing state-of-the-art explainability methods using 4 well-established evaluation metrics: Area Over the Perturbation Curve (AOPC)~\cite{eval-2}, Sensitivity~\cite{inf-sens}, Infidelity~\cite{inf-sens}, and Continuity~\cite{saifullah2022privacy}.
	
	\textbf{AOPC}~\cite{eval-2} measures the faithfulness of attributions by removing patches of the most relevant and least relevant image regions and subsequently computing the  area over the model's confidence curve for a given target class. 
	In this work, AOPC\textsubscript{MoRF} corresponds to the AOPC curves generated by removing the top most relevant features from the image, while AOPC\textsubscript{LeRF} corresponds to removing the least most relevant features. Higher values of AOPC\textsubscript{MoRF} and lower values of AOPC\textsubscript{LeRF} correspond to more faithful explanations. Another metric is the \textbf{Area Between the Perturbation Curves (ABPC)}~\cite{eval-2}, which summarizes the AOPC metric into a scalar value as the area between the AOPC\textsubscript{MoRF} and the AOPC\textsubscript{LeRF} curves. Larger ABPC values correspond to better explanations.  In this work, we compute the AOPC metrics by removing patches of size $8\times 8$ from the input samples using a pixel-flipping scheme where the white background regions are flipped to black and vice versa.
	
	\textbf{Sensitivity}~\cite{inf-sens} and \textbf{Infidelity}~\cite{inf-sens} both gauge the faithfulness and fidelity of explanations under input perturbations, with smaller values corresponding to better explanations for both of these metrics. 
	
	\textbf{Continuity}~\cite{saifullah2022privacy} is useful for measuring the overall interpretability of the attribution maps. It is computed by taking the mean of absolute gradients of the attribution map across horizontal and vertical directions. Smaller values for Continuity are desirable for better interpretability.
	
	To evaluate the metrics on the \rvlcdip{} dataset, we utilized 4000 randomly sampled images from the test set for all models, except EfficientNet-B4~\cite{effnet} and ConvNeXt-B~\cite{convnext}. For these two models, we limited the evaluation to 1000 test samples due to their extremely high computational requirements. For the Tobacco3482 dataset, we computed all metrics on the entire test set consisting of 700 image samples. In all experiments carried out in this work, we evaluate the attribution maps generated for the correctly predicted class.
	\begin{figure}[t!]
		\centering
		\includegraphics[width=0.8\textwidth]{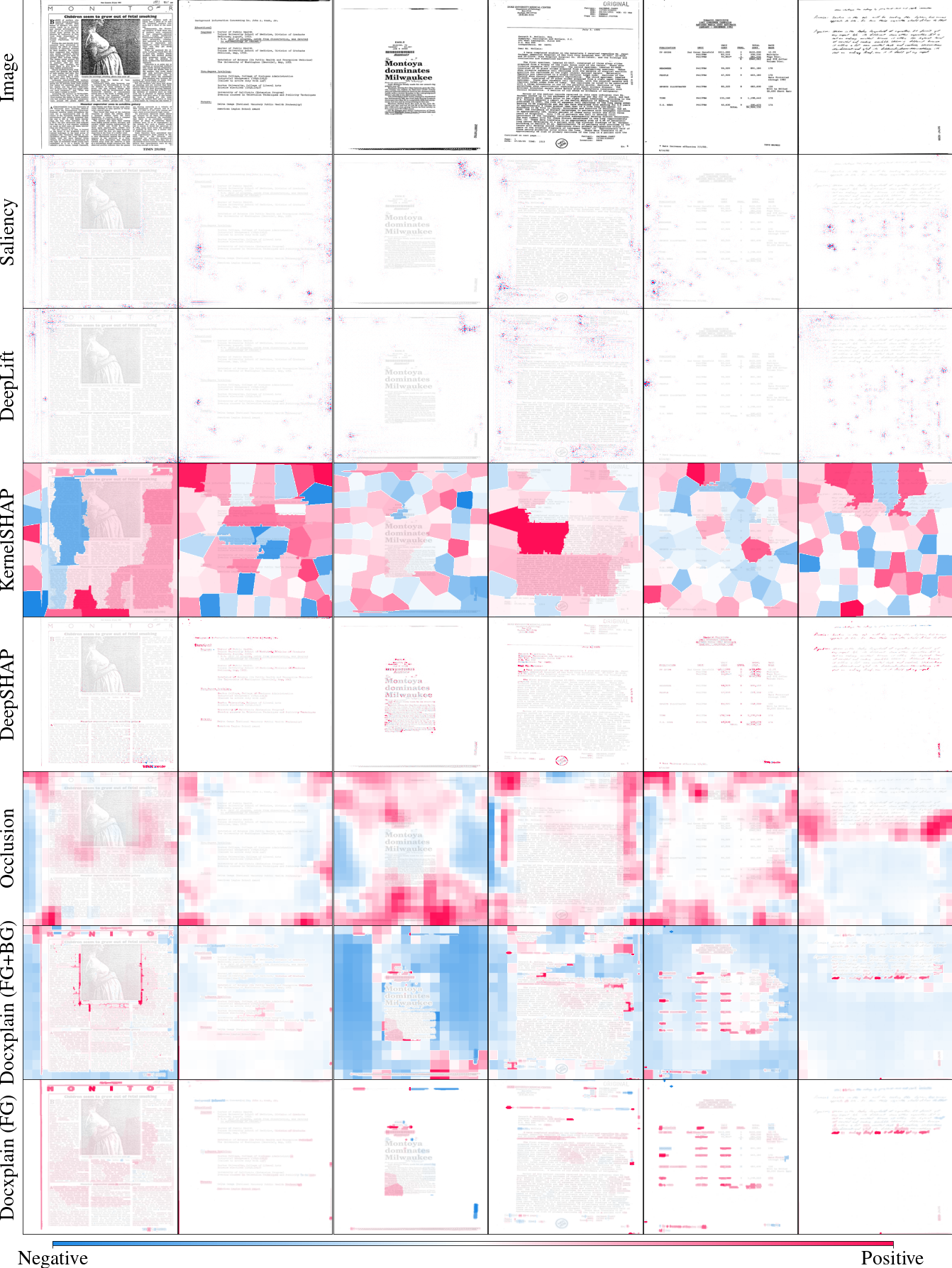}
		\caption{Explanations generated by different attribution methods for the ConvNeXt-B~\cite{convnext} model on 6 randomly selected samples from the RVL-CDIP dataset. As evident, our approach under both settings (\docxplainfg{} and \docxplainfgbg{}) produces significantly fine-grained attribution maps compared to existing methods. In addition, examining the two settings in combination allows decoupling whether an entire region or only specific foreground regions in the image are considered important by the model, significantly improving the interpretability of attributions.
			\vspace{-1em}
		} \label{fig:qualitative}
	\end{figure}
	
	\section{Experiment Results}
	In this section, we begin with a qualitative assessment of our proposed approach in comparison to existing attribution methods. Subsequently, we present the results of the quantitative evaluation of our approach using different evaluation metrics discussed in Section~\ref{sec:metrics}.
	\begin{figure}[t!]
		\centering
		\includegraphics[width=0.8\textwidth]{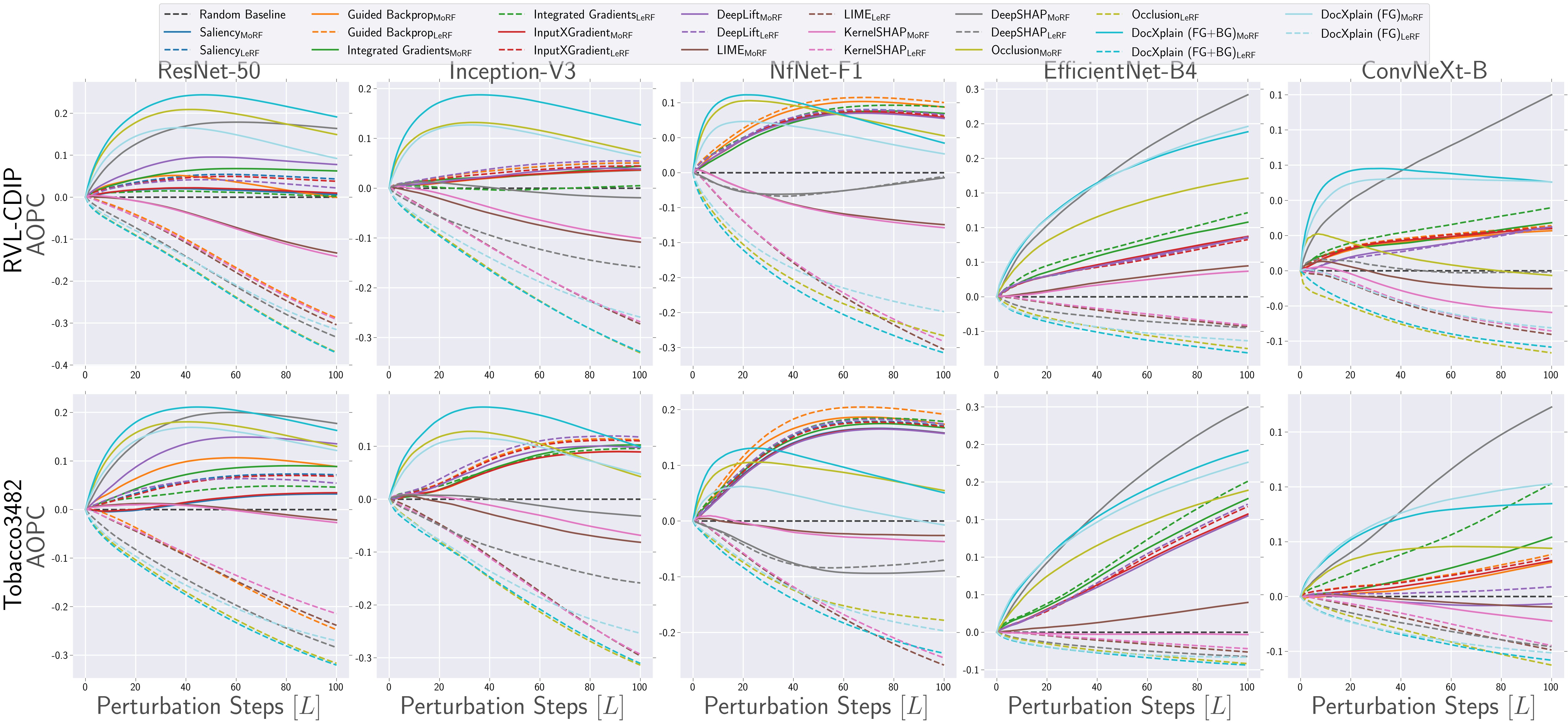}
		\caption{
			A comparison of AOPC\textsubscript{MoRF} and AOPC\textsubscript{LeRF} of different methods, plotted relative to the random baseline, across 5 selected deep neural networks. As evident from the steep rise and steep descent in the AOPC\textsubscript{MoRF} and AOPC\textsubscript{LeRF} curves of our approach under both settings, it demonstrates significantly high faithfulness compared to other methods.} \label{fig:aopc}
	\end{figure}
	
	\subsection{Qualitative Comparison} 
	\label{sec:qual}
	In Fig.~\ref{fig:qualitative}, we compare the explanations generated for the ConvNeXt-B~\cite{convnext} model by our proposed approach under two different settings, \docxplainfgbg{} and \docxplainfg{}, with those generated by 5 existing attribution methods on 6 randomly selected samples from the \rvlcdip{} dataset.	
	Several notable observations can be drawn from these results. Firstly, it is directly evident from the quality of attributions that both gradient-based approaches (\saliency{}, and \deeplift{}) and perturbation-based approaches (\kernelshap{}) perform significantly poorer compared to our proposed approach, especially under the \docxplainfg{} setting.	
	On the other hand, \occlusion{} exhibited noticeable similarities with our approach. Particularly, it can be observed that our approach generates structurally similar attribution maps to \occlusion{} when applied under the \docxplainfgbg{}, suggesting a consensus between our approach and \occlusion{} on the importance of regions in the image. 
	However, it is also noticeable that, compared to \occlusion{}, our approach produces significantly more interpretable maps, with concrete regions of importance, owing to the distinct boundaries between the foreground and background regions. Furthermore, examining the attribution maps of both \docxplainfg{} and \docxplainfgbg{} in combination allows distinguishing whether an entire region is important (including both foreground and background) or only a specific textual or image foreground feature in a given region holds significance. 	
	In contrast, when comparing \deepshap{} with the \docxplainfg{} setting, although the two approaches showed a slight visual resemblance, they diverged significantly on the importance of features. 
	While in some cases, they reached a consensus, in others they exhibited major discrepancies with opposite attributions assigned to some features. 	
	In addition, it can be observed that \deepshap{} generally produced highly sparse attribution maps, sometimes only lightly highlighting the boundaries of text strokes or individual letters. In contrast, our approach under the \docxplainfg{} setting assigns importance to complete words, textual lines, text blocks, and structural elements such as images and tables in the image, allowing for significantly better interpretability. Additional qualitative results across different models and methods are presented in Appendix~\ref{app:qual}.

	\begin{figure}[t!]
		\centering
		\includegraphics[width=0.8\textwidth]{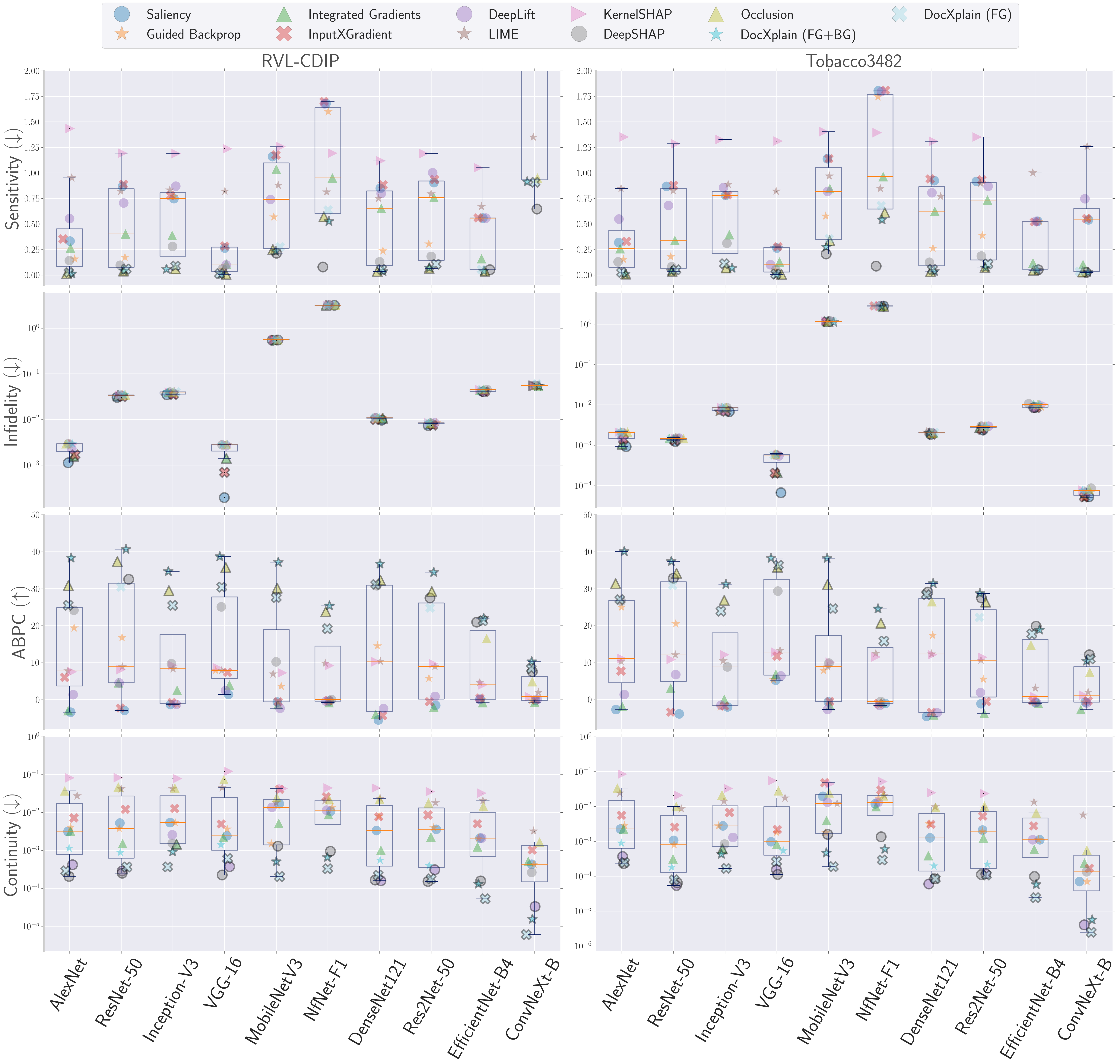}
		\caption{
			The results of 4 metrics, Sensitivity~\cite{inf-sens}, Infidelity~\cite{inf-sens}, Continuity~\cite{saifullah2022privacy}, and ABPC~\cite{eval-2} obtained by each method are shown for each model on the X-axis. It can be observed that our approach, under both settings, either outperforms or performs comparably to existing state-of-the-art attribution-based approaches on various explainability metrics. } \label{fig:quantitative}
	\end{figure}
	\subsection{Quantitative Comparison}
	\label{sec:quan}
	In Fig.~\ref{fig:aopc}, we compare the AOPCs of our proposed method (under both \docxplainfg{} and \docxplainfgbg{} settings) with existing attribution approaches on 5 different deep neural networks investigated in this work. 		
	Firstly, it can be observed that in terms of the AOPCs, our proposed approach demonstrated the best faithfulness across various models on both datasets, \rvlcdip{} and Tobacco3482. This is evident from the steep rise and steep descent in the AOPC\textsubscript{MoRF} and AOPC\textsubscript{LeRF} curves of our approach, respectively, in comparison to the other methods.	
	Another noteworthy observation is that, even under the \docxplainfg{} setting, where we only utilize the foreground features for generating explanations, our approach demonstrated significantly better faithfulness compared to other methods. 	
	In some cases, it performed similarly to \occlusion{}, as observed in the case of ResNet-50~\cite{resnet} and InceptionV3~\cite{inception} models, while in other cases, such as ConvNeX-B~\cite{convnext} and EfficientNet-B4~\cite{effnet}, it performed on a similar scale to \docxplainfgbg{} setting. 
	Furthermore, comparing the AOPCs of the two settings, \docxplainfgbg{} and \docxplainfg{}, provides valuable insight into whether a given model prioritizes only the foreground regions or focuses on both foreground and background for making a decision. Notably, for both ConvNeX-B~\cite{convnext} and EfficientNet-B4~\cite{effnet}, the AOPC\textsubscript{MoRF} curves for \docxplainfg{} and \docxplainfgbg{} closely align, indicating that these models assign relatively little importance to the background features in comparison to the other models. 	
	For additional comparison of AOPC on other models, see Appendix.~\ref{app:aopc}.
	
	In Fig.~\ref{fig:quantitative}, we present a quantitative evaluation of our approach against existing explainability methods, in terms of the Sensitivity~\cite{inf-sens}, Infidelity~\cite{inf-sens}, Continuity~\cite{saifullah2022privacy}, and the ABPC~\cite{eval-2} metrics. In particular, for each model on the X-axis, we make a box plot that summarizes the overall distribution of the metrics across the explainability methods and additionally plot the raw metric values for each attribution method for direct comparison.
	First, directing our attention to the Sensitivity~\cite{inf-sens} metric, it can be observed that our proposed approach under both settings performs equally well to the \occlusion{} and \deepshap{} approach, resulting in the lowest Sensitivity~\cite{inf-sens} across the majority of models. 
	Interestingly, on the NFNet-F1~\cite{nfnet} model, \deepshap{} scored considerably low on Sensitivity~\cite{inf-sens} compared to both \occlusion{} and our proposed approach. However, upon closer inspection of the corresponding ABPC scores, it becomes evident that \deepshap{}, along with many gradient-based approaches, performed extremely poorly on the ABPC metric on this model, indicating highly unfaithful explanations. This observation aligns with the findings of the previous study~\cite{Saifullah2023-docxai}, where \deepshap{} experienced failures on certain models, resulting in extremely poor attribution maps. Additional support for this observation is provided through supplementary qualitative results in Appendix~\ref{app:model-comparison}.		
	
	Focusing on the Infidelity~\cite{inf-sens} metric, it can be observed that all the approaches exhibited comparable and consistently low infidelity scores with minimal variance between the methods. However, on VGG-16~\cite{vgg} and AlexNet~\cite{alexnet}, the gradient-based approaches (especially Saliency~\cite{saliency}) appeared to perform slightly better on this metric.
	On the ABPC metric~\cite{eval-2}, which summarizes the results of the AOPC curves, we can observe that our approach greatly outperformed all existing approaches under the \docxplainfgbg{} setting, whereas \docxplainfg{} performed comparably to \deepshap{} and \occlusion{}, indicating better faithfulness. 
	Finally, on the Continuity metric~\cite{saifullah2022privacy}, which captures the interpretability of a method, our \docxplainfg{} approach consistently outperformed the majority of approaches and only occasionally fell behind \deeplift{} and \deepshap{}. 
	On the other hand, \docxplainfgbg{} lagged slightly behind the three methods (\deeplift{}, \deepshap{}, and \docxplainfg{}), indicating poorer interpretability in comparison. 	
	Note that this relatively poorer interpretability of \docxplainfgbg{} is also evident from the qualitative results in Fig.~\ref{fig:qualitative}, where, when observed independently of \docxplainfg{}, the interpretation of attribution maps was challenging. However, it is also worth noting that, despite the similarities in qualitative results with \occlusion{}, \docxplainfgbg{} still resulted in significantly lower Continuity~\cite{saifullah2022privacy} scores than the latter (notice the log scale). For a more detailed examination, the raw metric values for all the attribution methods are also provided in Appendix.~\ref{app:metrics}.
	\section{Conclusion}
	In this paper, we presented a novel model-agnostic explainability method especially designed for document image classification. Through a thorough qualitative and quantitative evaluation of our approach across different settings and various metrics, we demonstrated the superiority of our approach compared to existing methods in terms of both faithfulness and interpretability.	
	In addition, we demonstrated how our approach enables decoupling the foreground and background feature importance, enhancing the interpretation of attribution maps. Future efforts could focus on refining the segmentation approach by combining it with optical-character-recognition (OCR) models for direct text localization. In addition, it could be worthwhile to explore extending this approach to other document image analysis tasks and to multimodal document analysis models.

	\bibliographystyle{splncs04}

	\newpage
	\begin{subappendices}
		\renewcommand{\thesection}{\Alph{section}}%
		\section{Additional Qualitative Results}
		\label{app:qual}
		\subsection{Additional Qualitative Comparison of Methods}
\begin{figure}[h!]
			\centering
			\includegraphics[width=0.8\textwidth]{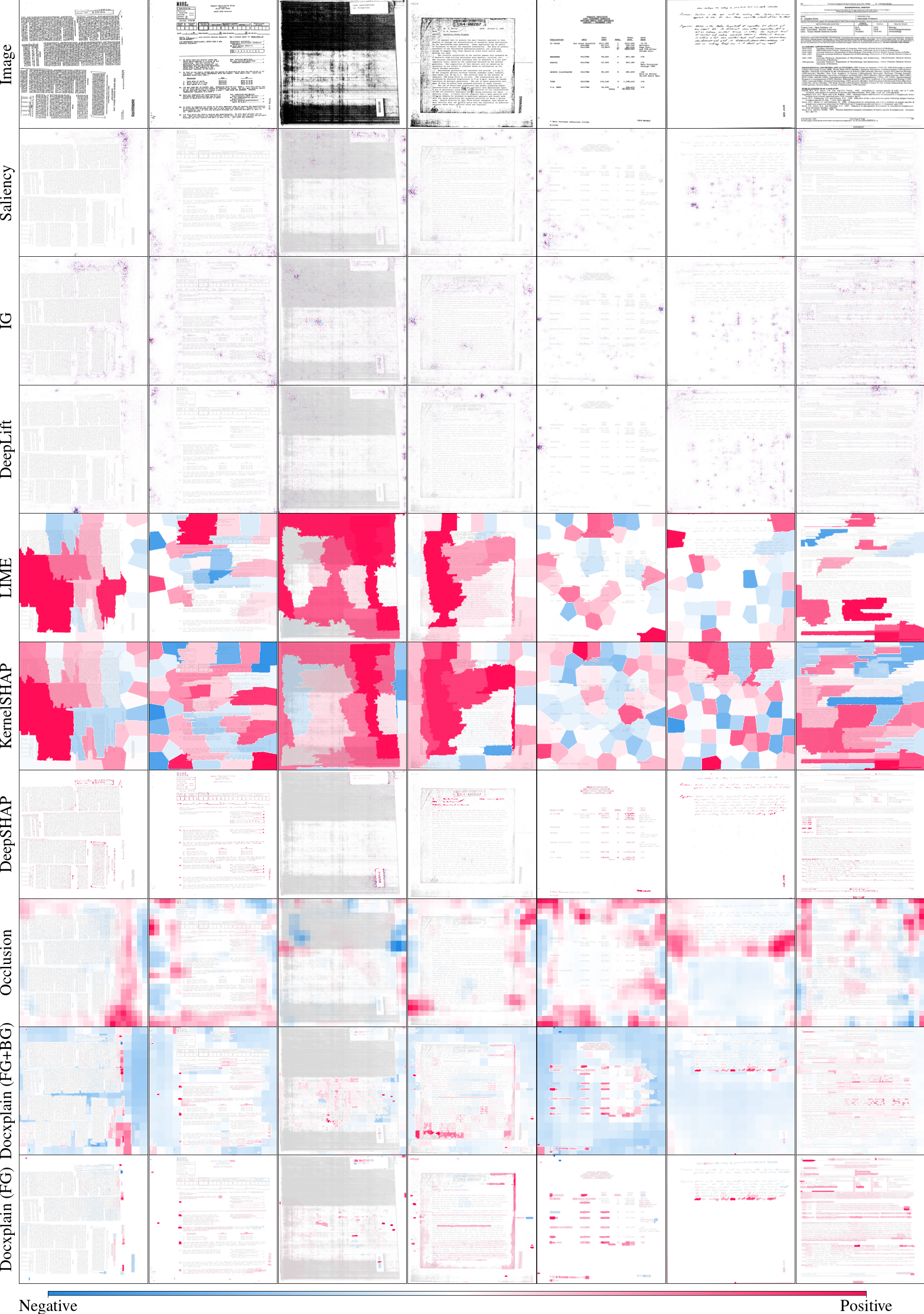}
			\caption{Additional comparison of explanations generated for the ConvNeXt-B~\cite{convnext} model by different attribution methods on 7 randomly selected samples from the \rvlcdip{} dataset.} \label{fig:app_add_results_convnext_base}
		\end{figure}

\begin{figure}[h!]
			\centering
			\includegraphics[width=0.8\textwidth]{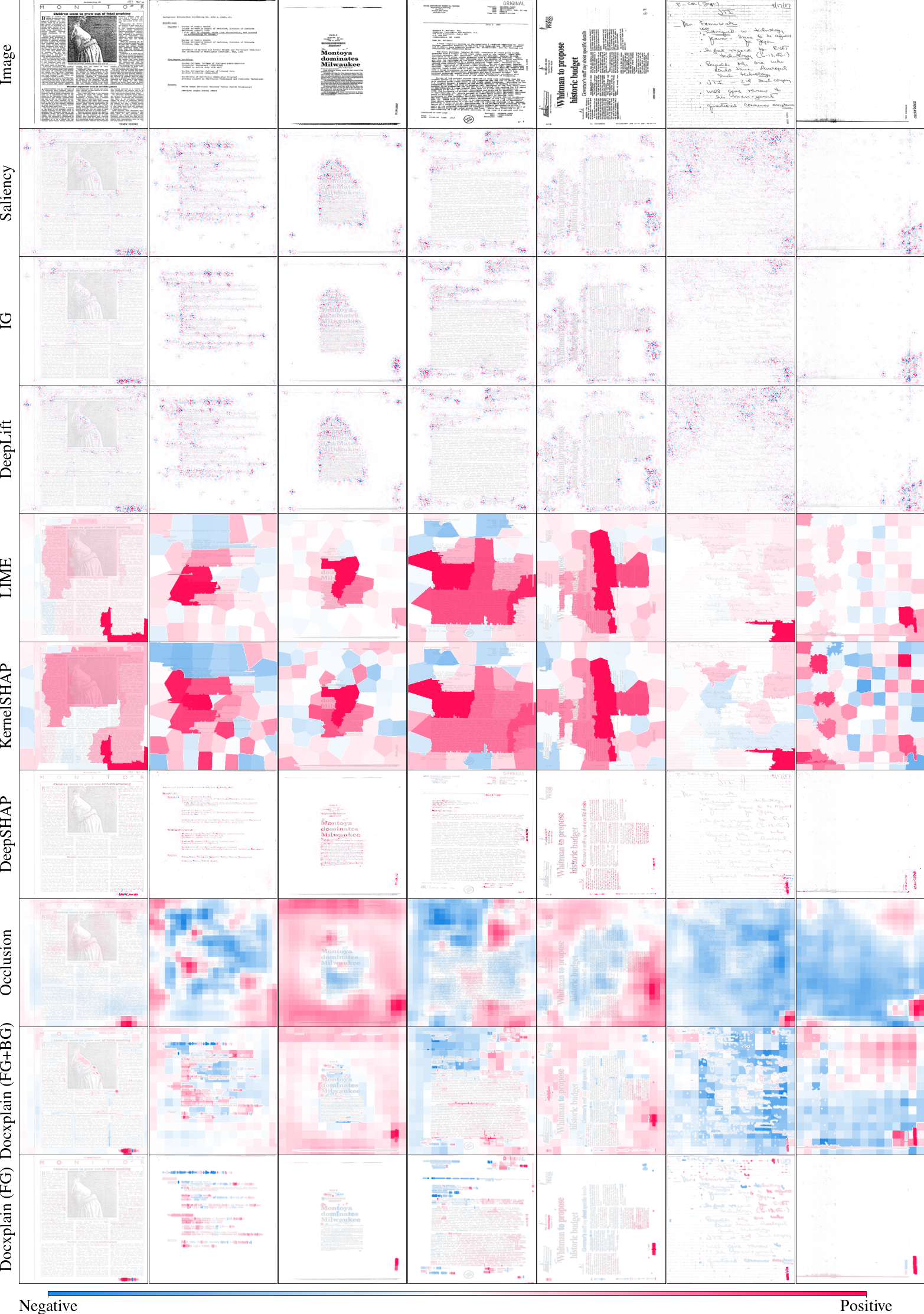}
			\caption{Additional comparison of explanations generated for the EfficientNet-B4~\cite{effnet} model by different attribution methods on 7 randomly selected samples from the \rvlcdip{} dataset.} \label{fig:app_add_results_eff}
		\end{figure}

\begin{figure}[h!]
			\centering
			\includegraphics[width=0.8\textwidth]{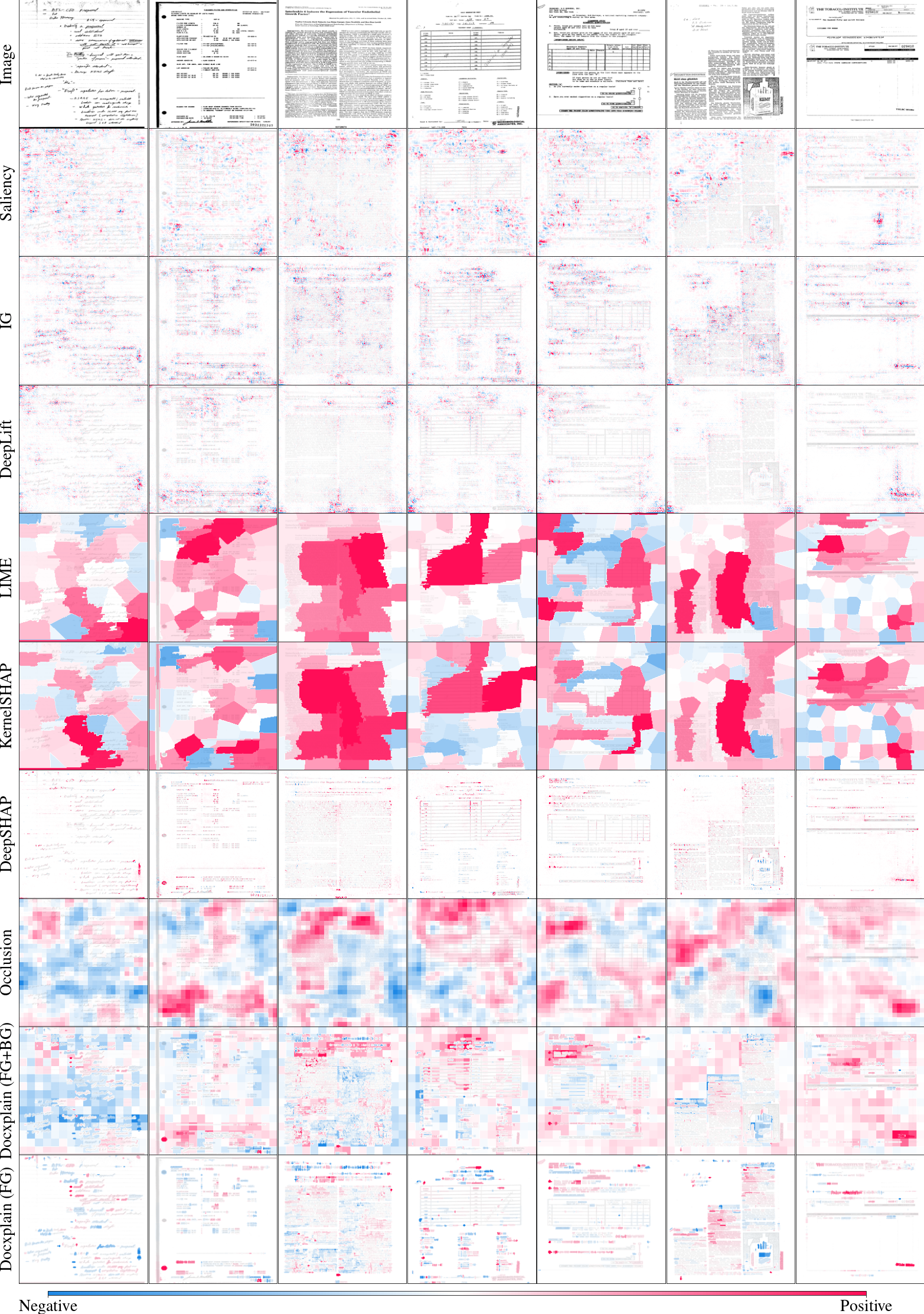}
			\caption{Additional comparison of explanations generated for the ResNet-50~\cite{resnet} model by different attribution methods on 7 randomly selected samples from the \rvlcdip{} dataset.} \label{fig:app_add_results_resnet50}
		\end{figure}

		In Fig.~\ref{fig:app_add_results_convnext_base}, Fig.~\ref{fig:app_add_results_eff}, and Fig.~\ref{fig:app_add_results_resnet50}, we show additional results for the explanations generated by different explainability methods for the ConvNeXt-B~\cite{convnext}, EfficientNet-B4~\cite{effnet}, and ResNet-50~\cite{resnet} models, respectively, on 6 random selected images from the \rvlcdip{} dataset.
		By examining the figures across the models, it can be noted that our observations from Section.~\ref{sec:qual} about the gradient-based approaches and perturbation-based approaches remain consistent across different models. Similarly, our observations regarding the similarities between \occlusion{} and \docxplainfgbg{}, as well as the diverging consensus between \docxplainfg{} and \deepshap{}, also remain consistent across different models.		
		\subsection{Additional Qualitative Comparison across Models}
\begin{figure}[h!]
			\centering
			\includegraphics[width=0.93\textwidth]{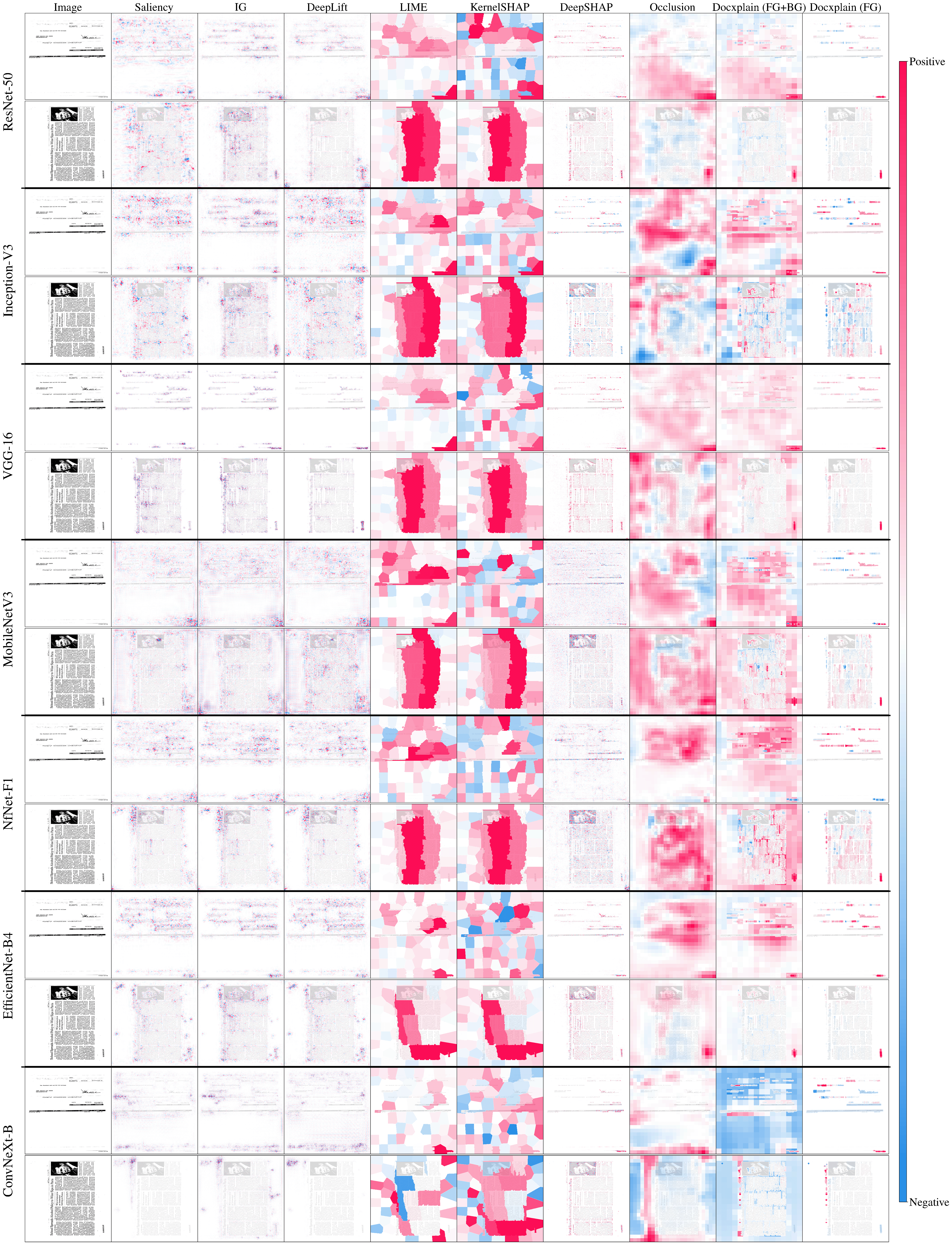}
			\caption{A comparison of explanations generated by different attribution methods for 7 different model on 2 randomly selected samples from the \rvlcdip{} dataset. It can be observed that the assessment of feature importance varies significantly across the models. In addition, it can be noticed that \deepshap{} fails to produce reasonable results on some models: InceptionV3~\cite{inception}, MobileNetV3~\cite{mobilenet}, and NFNet-F1~\cite{nfnet} } \label{fig:app_qualitative}
		\end{figure}

		Fig.~\ref{fig:app_qualitative} provides a qualitative comparison of the attribution methods across 7 different models on 2 random selected images from the \rvlcdip{} dataset. 
		
		\noindent First, it is evident that the attributions vary significantly across the models. However, a noticeable consensus can still be observed between \docxplainfgbg{} and \occlusion{} across different models. 
		Furthermore, it can be noted that \deepshap{} fails to produce reasonable attribution maps for the three models: InceptionV3~\cite{inception}, MobileNetV3~\cite{mobilenet}, and NFNet-F1~\cite{nfnet}. This behavior was observed on \deepshap{} across both various samples and datasets.
		\label{app:model-comparison}
		\section{Additional Quantitative Results}	
		\subsection{Additional AOPC Results}			
\begin{figure}[h!]
			\includegraphics[width=\textwidth]{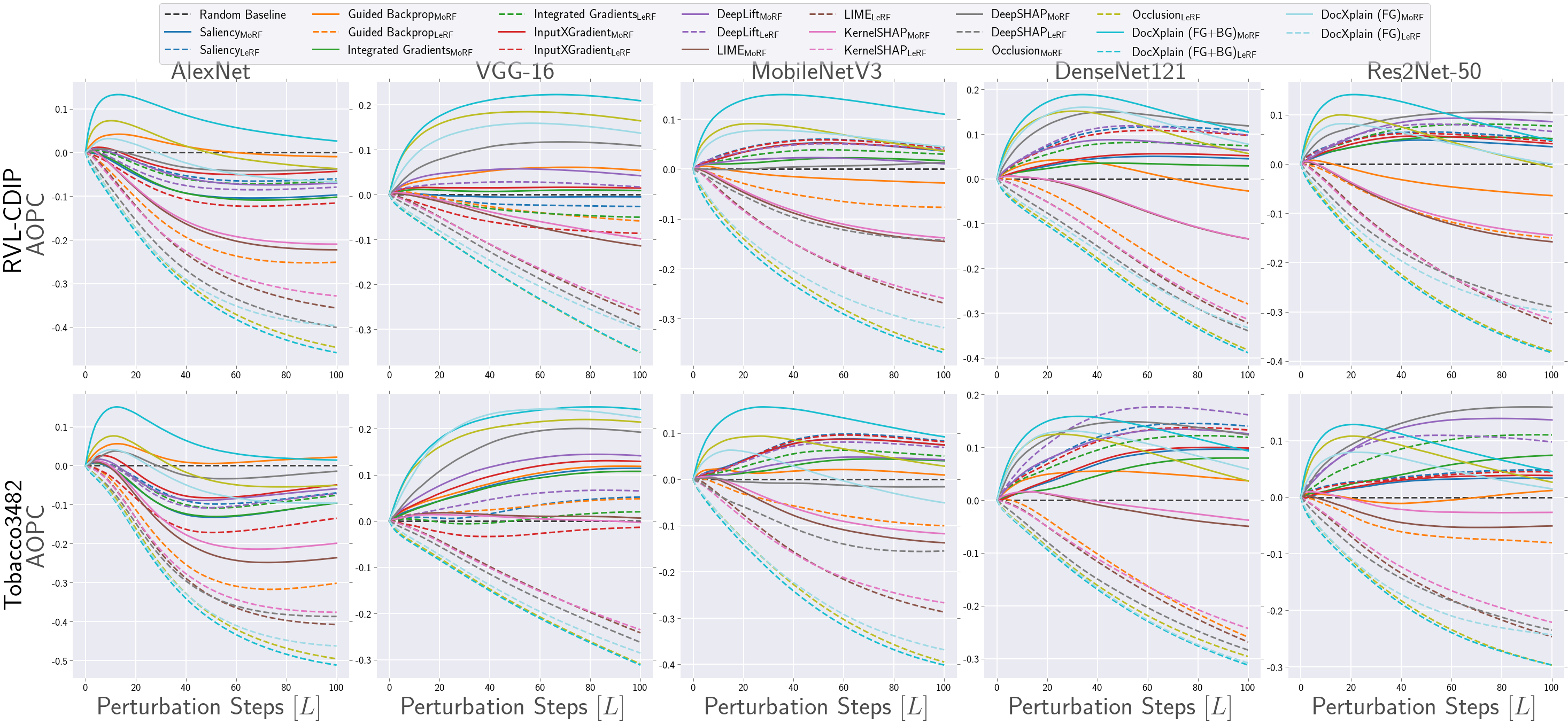}
			\caption{
				Additional AOPCs generated by the methods on rest of the deep neural networks. As evident, the behavior of the AOPC curves remains consistent across various models with our approach performing the best across both datasets.
			} \label{fig:aopc_app}
		\end{figure}

		In Fig.~\ref{fig:aopc_app}, we present additional AOPC results obtained for models not considered in the main text. It can be noted that our previous observations from Section.~\ref{sec:quan} about the faithfulness of different explainability methods are consistent across these models.
		\label{app:aopc}
		\newpage
		\subsection{Metric Results}
		\label{app:metrics}
		The raw metric values computed for all the models on the two datasets \rvlcdip{} and Tobacco3482 are given in Table.~\ref{tab:metrics}
		\begin{table}[h!]
			\caption{The metric values computed for each model across the two datasets \rvlcdip{} and Tobacco3482. For each metric, the top \firstval{1st}, \secondval{2nd}, and \thirdval{3rd} methods are bolded, italicized, and underlined, respectively}\label{tab:metrics}
			\centering
			\resizebox{0.85\textwidth}{!}{
				\begin{tabular}{|l|l|l|c|r|r|r|r|r|r|r|r|r|r|r|}
					\hline
					\setlength\extrarowheight{6pt}
					&Models&&&\multicolumn{11}{|c|}{Methods}\\\hline
					&&Acc.&&Saliency &	Guid. Back.&	IG	&InputXGradient	&DeepLift	&LIME	&KernalSHAP	&DeepSHAP&	Occclusion	&DocXplain\textsubscript{BG+FG}&	DocXplain\textsubscript{FG}\\
					\hline
					\multirow{40}{*}{\rotatebox{90}{RVL-CDIP}}
					&\multirow{4}{*}{AlexNet}&\multirow{4}{*}{87.86}&Sens. $\downarrow$&0.3326&0.1574&0.264&0.3537&0.5521&0.9521&1.4343&0.142&\secondval{0.0149}&\firstval{0.0121}&\thirdval{0.0259}\\\cline{4-15}
					&&&Inf. $\downarrow$&\firstval{0.0011}&0.0029&\secondval{0.0016}&\thirdval{0.0017}&0.0023&0.0029&0.0029&0.0029&0.003&0.0029&0.0029\\\cline{4-15}
					&&&Cont. $\downarrow$&0.31&0.41&0.32&0.72&\thirdval{0.04}&2.75&8.18&\firstval{0.02}&3.74&0.11&\secondval{0.03}\\\cline{4-15}
					&&&ABPC $\uparrow$&-3.41&19.38&-2.97&6.02&1.34&7.78&7.54&24.15&\secondval{30.82}&\firstval{38.31}&\thirdval{25.52}\\\cline{2-15}
					&\multirow{4}{*}{ResNet50}&\multirow{4}{*}{90.40}&Sens. $\downarrow$&0.869&0.1731&0.4025&0.89&0.7059&0.8205&1.1937&0.0959&\firstval{0.0394}&\secondval{0.0415}&\thirdval{0.0588}\\\cline{4-15}
					&&&Inf. $\downarrow$&\firstval{0.0305}&0.0344&\thirdval{0.0337}&\secondval{0.0312}&0.0338&0.0346&0.0341&0.0341&0.0343&0.0347&0.0347\\\cline{4-15}
					&&&Cont. $\downarrow$&0.53&0.38&0.15&1.22&\firstval{0.03}&4.23&8.31&\secondval{0.03}&4.73&0.09&\thirdval{0.04}\\\cline{4-15}
					&&&ABPC $\uparrow$&-2.93&16.79&4.55&-2.28&4.53&8.88&8.09&\thirdval{32.54}&\secondval{37.31}&\firstval{40.67}&30.45\\\cline{2-15}
					&\multirow{4}{*}{InceptionV3}&\multirow{4}{*}{91.18}&Sens. $\downarrow$&0.7484&0.7687&0.3887&0.7794&0.8701&0.8329&1.1889&0.2809&\secondval{0.0607}&\firstval{0.058}&\thirdval{0.0892}\\\cline{4-15}
					&&&Inf. $\downarrow$&\firstval{0.0349}&\secondval{0.0349}&0.0399&\thirdval{0.0351}&0.0372&0.0403&0.0399&0.0399&0.0402&0.0403&0.0402\\\cline{4-15}
					&&&Cont. $\downarrow$&0.54&0.54&\thirdval{0.15}&1.27&0.26&4.24&7.88&0.15&4.39&\secondval{0.09}&\firstval{0.04}\\\cline{4-15}
					&&&ABPC $\uparrow$&-1.34&-1.33&2.54&-0.75&-1.27&8.35&9.26&9.7&\secondval{29.45}&\firstval{34.67}&\thirdval{25.48}\\\cline{2-15}
					&\multirow{4}{*}{VGG-16}&\multirow{4}{*}{90.96}&Sens. $\downarrow$&0.2645&0.0546&0.1092&0.2839&0.0999&0.8213&1.2375&0.0592&\firstval{0.0067}&\secondval{0.007}&\thirdval{0.0138}\\\cline{4-15}
					&&&Inf. $\downarrow$&\firstval{0.0002}&0.0029&\thirdval{0.0014}&\secondval{0.0007}&0.0027&0.0028&0.0029&0.0028&0.0028&0.0028&0.0028\\\cline{4-15}
					&&&Cont. $\downarrow$&0.24&0.37&0.22&0.5&\secondval{0.04}&4.53&12.18&\firstval{0.02}&7.39&0.14&\thirdval{0.06}\\\cline{4-15}
					&&&ABPC $\uparrow$&1.4&7.9&3.96&7.39&2.47&7.9&8.65&25.08&\secondval{35.74}&\firstval{38.69}&\thirdval{30.43}\\\cline{2-15}
					&\multirow{4}{*}{MobileNetV3}&\multirow{4}{*}{87.56}&Sens. $\downarrow$&1.1594&0.5692&1.0371&1.175&0.7394&0.8798&1.2574&\firstval{0.2143}&\thirdval{0.25}&\secondval{0.2333}&0.2756\\\cline{4-15}
					&&&Inf. $\downarrow$&\firstval{0.544}&0.5626&0.5606&\secondval{0.5513}&\thirdval{0.5523}&0.562&0.5629&0.5649&0.5639&0.5616&0.5643\\\cline{4-15}
					&&&Cont. $\downarrow$&1.7&0.15&0.51&4.11&1.37&1.81&4.32&\thirdval{0.13}&2.55&\secondval{0.05}&\firstval{0.02}\\\cline{4-15}
					&&&ABPC $\uparrow$&-0.66&3.61&-1.34&-0.56&-2.35&6.93&7.24&10.21&\secondval{30.14}&\firstval{37.16}&\thirdval{27.6}\\\cline{2-15}
					&\multirow{4}{*}{NFNet-F1}&\multirow{4}{*}{88.83}&Sens. $\downarrow$&1.6795&1.6&0.9517&1.7007&1.6771&0.8136&1.194&\firstval{0.0791}&\thirdval{0.5711}&\secondval{0.5264}&0.6367\\\cline{4-15}
					&&&Inf. $\downarrow$&\secondval{3.1931}&\thirdval{3.195}&\firstval{3.1843}&3.2275&3.2206&3.2157&3.2055&3.2022&3.1954&3.2193&3.1983\\\cline{4-15}
					&&&Cont. $\downarrow$&1.08&1.38&0.88&2.59&1.15&2.04&4.46&\thirdval{0.1}&2.19&\secondval{0.07}&\firstval{0.03}\\\cline{4-15}
					&&&ABPC $\uparrow$&-0.36&-0.57&-0.8&-0.25&-0.46&9.81&9.22&0.01&\secondval{23.79}&\firstval{25.38}&\thirdval{19.17}\\\cline{2-15}
					&\multirow{4}{*}{DenseNet121}&\multirow{4}{*}{89.46}&Sens. $\downarrow$&0.8495&0.2371&0.6538&0.883&0.7981&0.7515&1.1191&0.1296&\firstval{0.0347}&\secondval{0.0358}&\thirdval{0.055}\\\cline{4-15}
					&&&Inf. $\downarrow$&\firstval{0.0097}&0.0109&\thirdval{0.0106}&\secondval{0.0099}&0.0108&0.0109&0.0109&0.0109&0.0109&0.0109&0.0109\\\cline{4-15}
					&&&Cont. $\downarrow$&0.33&0.8&0.1&0.77&\firstval{0.02}&2.36&4.43&\secondval{0.02}&2.26&0.06&\thirdval{0.02}\\\cline{4-15}
					&&&ABPC $\uparrow$&-5.51&14.52&-3.94&-4.34&-2.42&10.38&10.24&30.78&\secondval{32.28}&\firstval{36.7}&\thirdval{31.14}\\\cline{2-15}
					&\multirow{4}{*}{Res2Net50}&\multirow{4}{*}{88.96}&Sens. $\downarrow$&0.9056&0.3044&0.7598&0.9359&1.0016&0.795&1.1898&0.1829&\firstval{0.0654}&\secondval{0.0697}&\thirdval{0.1063}\\\cline{4-15}
					&&&Inf. $\downarrow$&\firstval{0.0074}&0.0085&\thirdval{0.0082}&\secondval{0.0075}&0.0085&0.0084&0.0084&0.0084&0.0084&0.0085&0.0084\\\cline{4-15}
					&&&Cont. $\downarrow$&0.36&0.4&0.23&0.86&\thirdval{0.03}&1.8&3.57&\firstval{0.02}&1.78&0.04&\secondval{0.02}\\\cline{4-15}
					&&&ABPC $\uparrow$&-1.56&5.76&-2.02&-0.57&0.85&8.96&9.7&\thirdval{27.48}&\secondval{29.23}&\firstval{34.45}&24.8\\\cline{2-15}
					&\multirow{4}{*}{EfficientNet-B4}&\multirow{4}{*}{92.68}&Sens. $\downarrow$&0.558&0.5586&0.1606&0.5598&0.5576&0.6702&1.052&\thirdval{0.0523}&\firstval{0.0382}&\secondval{0.039}&0.0546\\\cline{4-15}
					&&&Inf. $\downarrow$&0.0412&\thirdval{0.0408}&0.0435&\firstval{0.0403}&\secondval{0.0406}&0.0457&0.0452&0.0452&0.0463&0.0451&0.0462\\\cline{4-15}
					&&&Cont. $\downarrow$&0.21&0.21&0.12&0.5&0.21&2.05&3.24&\thirdval{0.02}&1.47&\firstval{0.01}&\secondval{0.01}\\\cline{4-15}
					&&&ABPC $\uparrow$&0.1&0.08&-0.81&0.39&0.06&4.64&4.02&\thirdval{20.96}&16.53&\firstval{22}&\secondval{21.32}\\\cline{2-15}
					&\multirow{4}{*}{ConvNeXt-B}&\multirow{4}{*}{93.89}&Sens. $\downarrow$&35.7846&35.7581&12.4655&36.4074&29.795&1.3507&\num{1.44e+06}&\firstval{0.6465}&0.9512&\thirdval{0.9139}&\secondval{0.9081}\\\cline{4-15}
					&&&Inf. $\downarrow$&0.0563&\firstval{0.0537}&0.0548&0.0558&0.0551&0.0562&\thirdval{0.0547}&0.0553&0.056&\secondval{0.0542}&0.0605\\\cline{4-15}
					&&&Cont. $\downarrow$&0.04&0.04&0.05&0.1&\firstval{0}&0.32&321470.51&0.03&0.17&\secondval{0}&\thirdval{0}\\\cline{4-15}
					&&&ABPC $\uparrow$&-0.27&-0.27&-0.71&-0.09&0.07&1.99&0.8&\thirdval{7.58}&4.86&\firstval{10.27}&\secondval{8.42}\\\hline
					
					\multirow{40}{*}{\rotatebox{90}{Tobacco3482}}
					&\multirow{4}{*}{AlexNet}&\multirow{4}{*}{88.85}&Sens. $\downarrow$&0.3184&0.1541&0.2587&0.3298&0.5484&0.8478&1.3525&0.1287&\firstval{0.0098}&\secondval{0.0103}&\thirdval{0.0258}\\\cline{4-15}
					&&&Inf. $\downarrow$&\firstval{0.0009}&0.0021&\secondval{0.0011}&\thirdval{0.0014}&0.0015&0.0022&0.0021&0.002&0.0021&0.0021&0.0021\\\cline{4-15}
					&&&Cont. $\downarrow$&0.23&0.29&0.23&0.56&\thirdval{0.04}&2.42&8.47&\firstval{0.02}&3.35&0.09&\secondval{0.02}\\\cline{4-15}
					&&&ABPC $\uparrow$&-2.69&25&-1.88&7.72&1.36&10.32&11.11&26.61&\secondval{31.36}&\firstval{40.07}&\thirdval{27.07}\\\cline{2-15}
					&\multirow{4}{*}{ResNet50}&\multirow{4}{*}{92.00}&Sens. $\downarrow$&0.8686&0.1805&0.3402&0.8786&0.6813&0.8262&1.2864&0.0821&\firstval{0.0365}&\secondval{0.0366}&\thirdval{0.0516}\\\cline{4-15}
					&&&Inf. $\downarrow$&\firstval{0.0012}&0.0015&0.0015&\secondval{0.0013}&0.0015&0.0016&0.0015&0.0015&0.0015&\thirdval{0.0014}&0.0015\\\cline{4-15}
					&&&Cont. $\downarrow$&0.11&0.08&0.03&0.26&\firstval{0.01}&0.86&2.1&\secondval{0.01}&0.99&0.02&\thirdval{0.01}\\\cline{4-15}
					&&&ABPC $\uparrow$&-3.86&20.52&3.12&-3.38&6.77&12.11&10.86&\thirdval{32.83}&\secondval{34.14}&\firstval{37.38}&30.88\\\cline{2-15}
					&\multirow{4}{*}{InceptionV3}&\multirow{4}{*}{92.71}&Sens. $\downarrow$&0.7801&0.7796&0.3957&0.784&0.8565&0.8882&1.3273&0.311&\secondval{0.0697}&\firstval{0.0682}&\thirdval{0.1106}\\\cline{4-15}
					&&&Inf. $\downarrow$&\firstval{0.0068}&\secondval{0.0068}&0.0082&\thirdval{0.0069}&0.0075&0.0087&0.0087&0.0086&0.0087&0.0087&0.0084\\\cline{4-15}
					&&&Cont. $\downarrow$&0.28&0.28&\thirdval{0.06}&0.67&0.13&1.41&3.28&0.08&2.13&\secondval{0.04}&\firstval{0.02}\\\cline{4-15}
					&&&ABPC $\uparrow$&-1.98&-1.98&0.1&-1.75&-1.51&10.51&12.16&8.85&\secondval{26.84}&\firstval{31.2}&\thirdval{23.92}\\\cline{2-15}
					&\multirow{4}{*}{VGG-16}&\multirow{4}{*}{93.71}&Sens. $\downarrow$&0.2643&0.0495&0.1279&0.2773&0.1013&0.8217&1.3114&0.0527&\secondval{0.0065}&\firstval{0.0062}&\thirdval{0.0096}\\\cline{4-15}
					&&&Inf. $\downarrow$&\firstval{0.0001}&0.0006&\secondval{0.0002}&\thirdval{0.0002}&0.0005&0.0006&0.0006&0.0006&0.0006&0.0006&0.0006\\\cline{4-15}
					&&&Cont. $\downarrow$&0.1&0.18&0.08&0.22&\secondval{0.02}&1.75&5.5&\firstval{0.01}&2.82&0.05&\thirdval{0.03}\\\cline{4-15}
					&&&ABPC $\uparrow$&5.24&5.42&6.89&11.77&6.4&13.32&12.85&29.34&\thirdval{35.8}&\firstval{38.26}&\secondval{36.38}\\\cline{2-15}
					&\multirow{4}{*}{MobileNetV3}&\multirow{4}{*}{85.85}&Sens. $\downarrow$&1.1391&0.5775&0.8473&1.1402&0.8186&0.9698&1.4042&\firstval{0.2065}&\thirdval{0.3363}&\secondval{0.2756}&0.3567\\\cline{4-15}
					&&&Inf. $\downarrow$&1.1612&1.1745&1.1575&\secondval{1.1561}&1.1764&1.1914&1.1899&1.1949&\thirdval{1.1562}&\firstval{1.1268}&1.1858\\\cline{4-15}
					&&&Cont. $\downarrow$&1.95&0.18&0.4&4.8&1.32&1.21&4.45&\thirdval{0.16}&2.47&\secondval{0.05}&\firstval{0.02}\\\cline{4-15}
					&&&ABPC $\uparrow$&-0.6&7.87&-1.61&-0.5&-2.67&8.92&10.11&9.88&\secondval{31.12}&\firstval{38.22}&\thirdval{24.62}\\\cline{2-15}
					&\multirow{4}{*}{NFNet-F1}&\multirow{4}{*}{91.28}&Sens. $\downarrow$&1.8044&1.7458&0.9637&1.8107&1.7957&0.8477&1.3942&\firstval{0.0877}&\thirdval{0.6119}&\secondval{0.5418}&0.6809\\\cline{4-15}
					&&&Inf. $\downarrow$&\thirdval{2.8069}&2.9079&2.8477&2.8308&2.9079&2.8397&2.8206&2.8506&\firstval{2.7411}&\secondval{2.7536}&2.8191\\\cline{4-15}
					&&&Cont. $\downarrow$&1.19&1.74&0.99&2.93&1.32&2.06&5.19&\thirdval{0.14}&2.03&\secondval{0.06}&\firstval{0.03}\\\cline{4-15}
					&&&ABPC $\uparrow$&-1.04&-1.47&-0.82&-1.03&-1.57&12.49&11.55&-0.51&\secondval{20.65}&\firstval{24.56}&\thirdval{15.83}\\\cline{2-15}
					&\multirow{4}{*}{DenseNet121}&\multirow{4}{*}{92.14}&Sens. $\downarrow$&0.924&0.2618&0.6259&0.9411&0.8062&0.7702&1.3101&0.1253&\firstval{0.0332}&\secondval{0.0363}&\thirdval{0.0561}\\\cline{4-15}
					&&&Inf. $\downarrow$&\firstval{0.0019}&0.0021&\thirdval{0.002}&\secondval{0.0019}&0.0021&0.0021&0.002&0.0021&0.002&0.002&0.0021\\\cline{4-15}
					&&&Cont. $\downarrow$&0.13&0.32&0.04&0.31&\firstval{0.01}&0.97&2.48&\secondval{0.01}&0.94&0.02&\thirdval{0.01}\\\cline{4-15}
					&&&ABPC $\uparrow$&-4.52&17.38&-4.12&-3.44&-3.56&12.34&12.13&\secondval{29.08}&26.49&\firstval{31.42}&\thirdval{28.37}\\\cline{2-15}
					&\multirow{4}{*}{Res2Net50}&\multirow{4}{*}{89.85}&Sens. $\downarrow$&0.9173&0.3873&0.7344&0.9325&0.8688&0.8984&1.3515&0.1863&\secondval{0.0738}&\firstval{0.0709}&\thirdval{0.1089}\\\cline{4-15}
					&&&Inf. $\downarrow$&\firstval{0.0024}&0.0029&\thirdval{0.0027}&\secondval{0.0025}&0.003&0.003&0.0029&0.003&0.003&0.003&0.0029\\\cline{4-15}
					&&&Cont. $\downarrow$&0.21&0.2&0.13&0.53&\firstval{0.01}&0.9&2.35&\secondval{0.01}&1.04&0.02&\thirdval{0.01}\\\cline{4-15}
					&&&ABPC $\uparrow$&-1.11&5.53&-3.72&-0.53&1.92&10.64&11.47&\secondval{27.5}&\thirdval{26.39}&\firstval{28.71}&22.24\\\cline{2-15}
					&\multirow{4}{*}{EfficientNet-B4}&\multirow{4}{*}{93.85}&Sens. $\downarrow$&0.5267&0.5264&0.1185&0.5201&0.5266&1.0015&\num{4.95e+06}&\thirdval{0.0515}&\firstval{0.0465}&\secondval{0.0486}&0.0628\\\cline{4-15}
					&&&Inf. $\downarrow$&\thirdval{0.0087}&0.0088&0.0102&\firstval{0.0085}&\secondval{0.0085}&0.0104&0.0103&0.0106&0.0106&0.0105&0.01\\\cline{4-15}
					&&&Cont. $\downarrow$&0.11&0.11&0.06&0.28&0.11&1.33&16059025&\secondval{0.01}&0.66&\thirdval{0.01}&\firstval{0}\\\cline{4-15}
					&&&ABPC $\uparrow$&-0.79&-0.79&-1.03&-0.3&-0.8&3.11&0.87&\firstval{19.85}&14.73&\secondval{18.86}&\thirdval{17.74}\\\cline{2-15}
					&\multirow{4}{*}{ConvNeXt-B}&\multirow{4}{*}{94.71}&Sens. $\downarrow$&0.5407&0.5562&0.1044&0.5527&0.7473&1.259&\num{3.10e+05}&\firstval{0.0255}&\thirdval{0.0303}&\secondval{0.0261}&0.0363\\\cline{4-15}
					&&&Inf. $\downarrow$&\firstval{0.0001}&\secondval{0.0001}&\thirdval{0.0001}&0.0001&0.0001&0.0001&0.0001&0.0001&0.0001&0.0001&0.0001\\\cline{4-15}
					&&&Cont. $\downarrow$&0.01&0.01&0.02&0.02&\firstval{0}&0.57&594121.34&0.01&0.06&\secondval{0}&\thirdval{0}\\\cline{4-15}
					&&&ABPC $\uparrow$&-0.66&-0.66&-2.7&-0.32&-0.9&2.01&1.19&\firstval{12.13}&7.36&\thirdval{10.44}&\secondval{10.99}\\\hline
				\end{tabular}
			}
		\end{table}	
		\newpage	
	\end{subappendices}
\end{document}